\RequirePackage{fix-cm}
\documentclass[twocolumn]{svjour3}          
\smartqed  
\usepackage{graphicx}
\usepackage{times}
\usepackage{epsfig}
\usepackage{epstopdf}
\usepackage{amsmath}
\usepackage{amssymb}
\usepackage[labelfont=bf]{caption}
\usepackage{tabularx}
\usepackage[table]{xcolor}
\usepackage{algorithm}
\usepackage[noend]{algpseudocode}

\usepackage[section]{placeins}
\usepackage{booktabs}
\usepackage{multirow}
\usepackage{mmstyles}
\usepackage{siunitx}
\usepackage{marvosym}
\usepackage{natbib}
\definecolor{citecolor}{HTML}{0071bc}
\definecolor{tabhighlight}{HTML}{e5e5e5}
\usepackage[colorlinks,citecolor=citecolor]{hyperref}

\graphicspath{{figures/}}
\DeclareGraphicsExtensions{.pdf}
\makeatletter
\def\BState{\State\hskip-\ALG@thistlm}
\makeatother
\definecolor{myGreen}{HTML}{33FF00}
\definecolor{myRed}{HTML}{FF3030}
\definecolor{myGrey}{HTML}{AA5555}
\definecolor{myWhite}{HTML}{FFFFFF}
\definecolor{maroon}{cmyk}{0,0.87,0.68,0.32}
\definecolor{petr}{HTML}{5555FF}
\definecolor{josef}{HTML}{FF3030}

\journalname{IJCV}
\begin{document}
\begin{sloppypar}

\title{Delving into Inter-Image Invariance for Unsupervised Visual Representations}


\author{Jiahao Xie      \and
        Xiaohang Zhan   \and
        Ziwei Liu       \and
        Yew Soon Ong    \and
        Chen Change Loy
}


\institute{Jiahao Xie \at
              Nanyang Technological University \\
              \email{jiahao003@ntu.edu.sg}
           \and
           Xiaohang Zhan \at
              The Chinese University of Hong Kong \\
              \email{xiaohangzhan@outlook.com}
           \and
           Ziwei Liu \at
              Nanyang Technological University \\
              \email{ziwei.liu@ntu.edu.sg}
           \and
           Yew Soon Ong \at
              Nanyang Technological University \\
              \email{asysong@ntu.edu.sg}
           \and
           Chen Change Loy \at
              Nanyang Technological University \\
              \email{ccloy@ntu.edu.sg}
}
\date{Received: date / Accepted: date}

\maketitle

\begin{abstract}
Contrastive learning has recently shown immense potential in unsupervised visual representation learning. Existing studies in this track mainly focus on intra-image invariance learning. The learning typically uses rich intra-image transformations to construct positive pairs and then maximizes agreement using a contrastive loss.
The merits of inter-image invariance, conversely, remain much less explored. One major obstacle to exploit inter-image invariance is that it is unclear how to reliably construct inter-image positive pairs, and further derive effective supervision from them since no pair annotations are available. In this work, we present a comprehensive empirical study to better understand the role of inter-image invariance learning from three main constituting components: pseudo-label maintenance, sampling strategy, and decision boundary design. To facilitate the study, we introduce a unified and generic framework that supports the integration of unsupervised intra- and inter-image invariance learning. Through carefully-designed comparisons and analysis, multiple valuable observations are revealed: 1) online labels converge faster and perform better than offline labels; 2) semi-hard negative samples are more reliable and unbiased than hard negative samples; 3) a less stringent decision boundary is more favorable for inter-image invariance learning. With all the obtained recipes, our final model, namely InterCLR, shows consistent improvements over state-of-the-art intra-image invariance learning methods on multiple standard benchmarks. We hope this work will provide useful experience for devising effective unsupervised inter-image invariance learning. Code: \url{https://github.com/open-mmlab/mmselfsup}.

\keywords{Unsupervised learning \and Self-supervised learning \and Representation learning \and Contrastive learning \and Inter-image invariance}
\end{abstract}

\section{Introduction}
\label{sec:intro}

Unsupervised representation learning witnesses substantial progress thanks to the emergence of self-supervised learning\footnote{Self-supervised learning is a form of unsupervised learning. While these terms are used interchangeably in the literature, we use the more classical term of ``unsupervised learning'', to reflect the general sense of ``not supervised by human-annotated labels''. A more detailed treatment is provided in~\cite{ericsson2022self}.}, which can be broadly divided into four categories: recovery-based~\citep{doersch2015unsupervised,noroozi2016unsupervised,larsson2016learning,zhang2016colorful,pathak2016context,zhan2019self}, transformation prediction~\citep{dosovitskiy2014discriminative,liu2017video,gidaris2018unsupervised,zhang2019aet}, clustering-based~\citep{huang2016unsupervised,xie2016unsupervised,yang2016joint,caron2018deep,caron2019unsupervised,asano2020self,zhan2020online,caron2020unsupervised}, and contrastive learning~\citep{oord2018representation,wu2018unsupervised,tian2020contrastive,hjelm2019learning,ye2019unsupervised,zhuang2019local,he2020momentum,misra2020self,chen2020simple,grill2020bootstrap}.
Among the various paradigms, contrastive learning shows great potential and even surpasses supervised learning~\citep{he2020momentum,chen2020simple,grill2020bootstrap}.
A typical contrastive learning method applies rich transformations to an image and maximizes agreement between different transformed views of the same image via a contrastive loss in the latent feature space.
This process encourages the network to learn ``intra-image'' invariance (\ie, instance discrimination~\citep{wu2018unsupervised}). 

\begin{figure}[t]
	\centering
	\includegraphics[width=.99\linewidth]{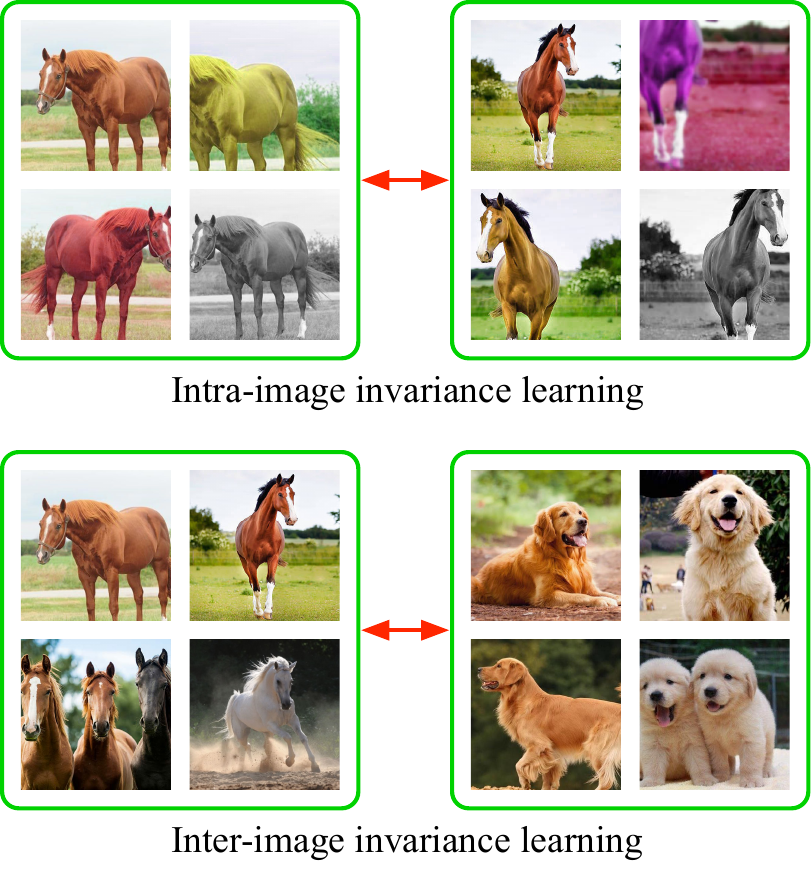}
	\caption{Intra-image invariance learning groups different augmented views of the same image together while separates different images apart. In contrast, inter-image invariance learning groups similar images together while separates dissimilar ones apart}
	\label{fig:teaser}
\end{figure}

Some typical ``intra-image'' transformations, including random cropping, resizing, flipping and color distortion, are shown in Fig.~\ref{fig:teaser}. 
Clearly, it is challenging to design convincing transformations to faithfully cover all the natural variances existing in natural images.
Hence, it remains an open question whether the existing form of transformations can sufficiently lead to our ideal representations, which should be invariant to viewpoints, occlusions, poses, instance-level or subclass-level differences.
Such variances naturally exist between pairs of instances belonging to the same semantic class.  
However, it is challenging to exploit such ``inter-image'' invariance in the context of unsupervised learning since no pair annotations are available.
Clustering is a plausible solution to derive such pseudo-labels for contrastive learning. For instance, LA~\citep{zhuang2019local} adopted off-the-shelf clustering to obtain pseudo-labels to constitute ``inter-image'' candidates.
Nevertheless, the performance still falls behind state-of-the-art intra-image invariance learning methods.
We believe there exist details that might have been ignored, if resolved, shall make the usefulness of inter-image invariance learning more pronounced than it currently does.

In this work, we go back to basics and systematically investigate the effects of inter-image invariance from three major aspects:

\noindent
\textbf{1) Pseudo-label maintenance}.
Owing to expensive computational cost, global clustering adopted in prior works~\citep{caron2018deep,zhuang2019local,li2021prototypical} can only be performed sparsely every several training epochs.
Hence, it inevitably produces stale labels relative to the rapidly updated network.
To re-assign pseudo-labels continuously and instantly, we consider mini-batch $k$-means in place of global $k$-means by integrating the label and centroid update steps into each training iteration.
In this way, clustering and network update are simultaneously undertaken, yielding more reliable pseudo-labels.

\noindent
\textbf{2) Sampling strategy}.
It is common for supervised learning to adopt hard negative mining~\citep{schroff2015facenet,oh2016deep,harwood2017smart,wu2017sampling,ge2018deep,suh2019stochastic}.
However, in the scenario of unsupervised learning, hard negatives might well have wrong labels, \ie, they may be actually positive pairs.
On the other hand, if we choose easy negative pairs and push them apart, they will still be easy negatives next time, and might never be corrected, leading to a shortcut solution.
Hence, the sampling strategy in unsupervised inter-image invariance learning is non-trivial, which has so far been neglected.

\noindent
\textbf{3) Decision boundary design}.
Existing works~\citep{liu2017sphereface,liu2016large,wang2018cosface,wang2018additive,deng2019arcface} in supervised learning design large-margin loss functions to learn discriminative features.
While in unsupervised learning, it is unsure whether pursuing discriminative features benefits since pseudo-labels are noisy.
For example, if a positive pair of images are misclassified as a negative one, the large-margin optimization strategy will further push them apart. Then the situation will never be corrected.
We explore decision margin designs for both the unsupervised intra- and inter-image branches.

\textbf{Contributions} -- This study aims at revealing key aspects that should be carefully considered when leveraging unsupervised inter-image information in contrastive learning. 
Although some of these aspects were originally considered in supervised learning, the conclusions are quite different and unique in the unsupervised scenario. To the best of our knowledge, this is the first empirical study on the effects of these aspects for unsupervised inter-image contrastive representations.
The merits of inter-image invariance learning are demonstrated through its consistent improvements over state-of-the-art intra-image invariance learning methods on multiple standard benchmarks.
%

\section{Related Work} 
\label{sec:work}

\noindent\textbf{Contrastive-based representation learning.}
Contrastive-based methods learn invariant features by contrasting positive samples against negative ones. A positive pair is usually formed with two augmented views of the same image, while negative ones are formed with different images. Typically, the positive and negative samples can be obtained either within a batch or from a memory bank. In batch-wise methods~\citep{oord2018representation,hjelm2019learning,ye2019unsupervised,henaff2019data,bachman2019learning,chen2020simple}, positive and negative samples are drawn from the current mini-batch with the same encoder that is updated end-to-end with back-propagation.
For methods based on memory bank~\citep{wu2018unsupervised,tian2020contrastive,zhuang2019local,misra2020self}, positive and negative samples are drawn from a memory bank that stores features of all samples computed in previous steps. Recently, MoCo~\citep{he2020momentum} builds large and consistent dictionaries for contrastive learning using a slowly progressing encoder. BYOL~\citep{grill2020bootstrap} and SimSiam~\citep{chen2021exploring} further learn invariant features without negative samples.
As opposed to our work, the aforementioned approaches only explore intra-image statistics for contrastive learning.
Although there are a few prior attempts~\citep{zhuang2019local,li2021prototypical} to leverage inter-image statistics for contrastive learning, they mainly focus on either designing sampling metric or comparing instance-group features while leaving other important aspects unexplored. NNCLR~\citep{dwibedi2021little} also embraces inter-image samples for contrastive learning. As opposed to our work, they use nearest neighbors from a support set to define positive samples, whereas we use cluster assignments to sample contrastive pairs. Besides, NNCLR still relies on a large batch size (\eg, 4096) to achieve good results, while using a smaller batch size (\eg, 256) will significantly decrease its performance. In contrast, InterCLR can achieve competitive performance using a more affordable batch size of 256. More importantly, we empirically study inter-image invariance learning from different aspects and show that pseudo-label maintenance, sampling strategy and decision boundary design should be collectively considered for good results.

Our study is more related to a strand of recent research~\citep{saunshi2019theoretical,wang2020understanding,tian2020makes,purushwalkam2020demystifying,tosh2021contrastive,zhao2021makes,xiao2021should} that focuses on developing theoretical or empirical understanding of contrastive representations from different aspects.
As opposed to their works, we provide empirical understanding of contrastive learning from its inter-image invariance perspective.

There are also another group of works that perform dense contrastive learning by extending existing image-level methods to the pixel level~\citep{pinheiro2020unsupervised,wang2021dense,xie2021propagate,selvaraju2021casting,liu2020self,henaff2021efficient} or the region level~\citep{roh2021spatially,yang2021instance,xiao2021region,xie2021detco,ding2021unsupervised,xie2021unsupervised,wei2021aligning}. Although better performance emerges on dense prediction downstream tasks, most of their classification downstream performance is largely sacrificed. Our work differs from this line of research in that we do not aim at developing a more advanced pretext task to learn spatially structured representations. Instead, we aim at better leveraging inter-image invariance for contrastive learning to pursue generic representations that improve both classification and dense prediction downstream tasks. We expect that our findings can be further applied on these dense contrastive learning variants.

\begin{figure*}[t]
	\centering
	\includegraphics[width=\linewidth]{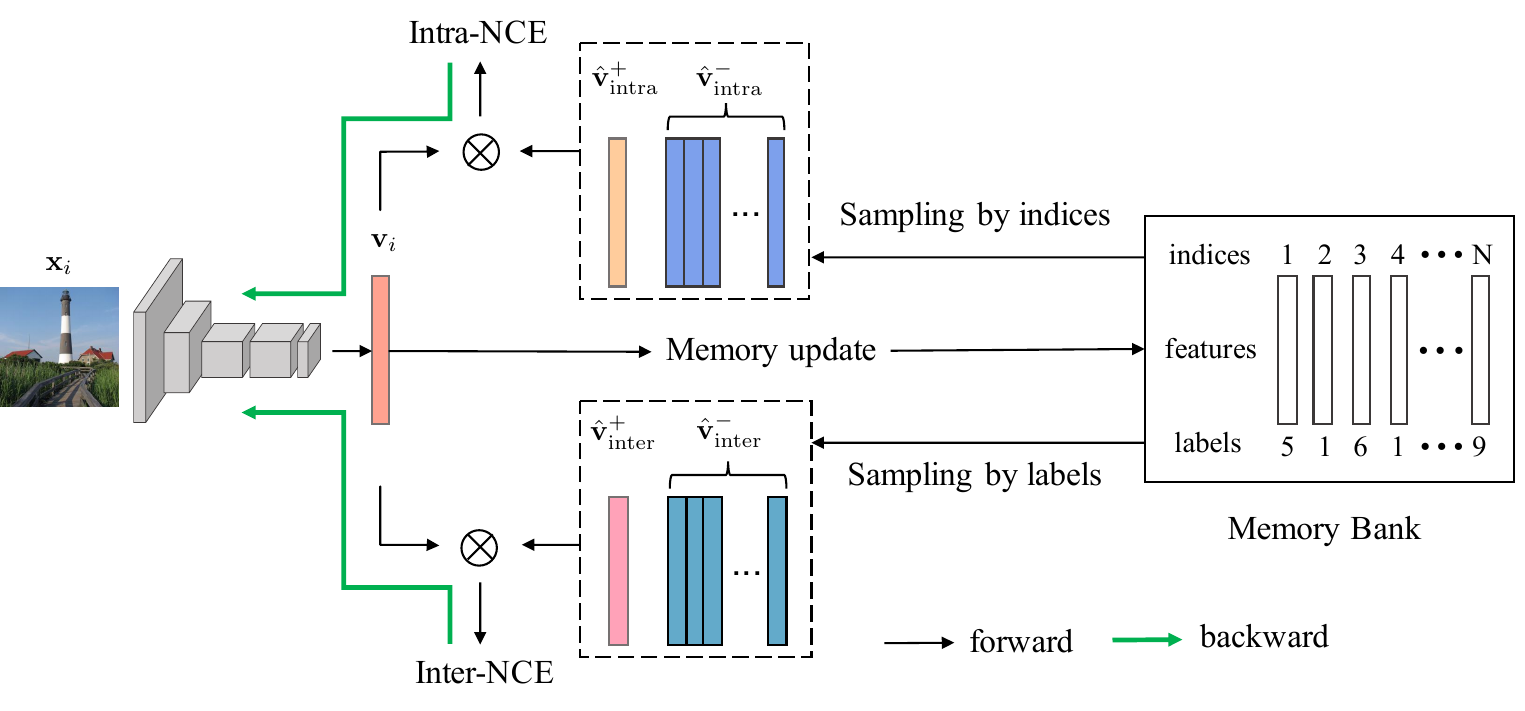}
	\caption{\textbf{Overview of our unified intra- and inter-image invariance learning framework (InterCLR).} For the intra-image component, positive and negative pairs are sampled by indices, while for the inter-image part, they are sampled by pseudo-labels. The memory bank including features and pseudo-labels is updated in each iteration. ``Intra-NCE'' and ``Inter-NCE'', the variants of InfoNCE~\citep{oord2018representation}, constitute loss functions for the two branches, respectively}
	\label{fig:framework}
\end{figure*}

\noindent\textbf{Clustering-based representation learning.}
Earlier attempts have shown great potential of joint clustering and feature learning, but the studies are limited to small datasets~\citep{xie2016unsupervised,yang2016joint,liao2016learning,bojanowski2017unsupervised,chang2017deep,ji2019invariant}. DeepCluster~\citep{caron2018deep} (DC) scales up the learning to millions of images through alternating between deep feature clustering and CNN parameters update. Although DC uses clustering during representation learning, it differs from our work conceptually in two aspects. First, DC optimizes the cross-entropy loss between predictions and pseudo-labels obtained by cluster assignments. Such optimization requires an additional parametric classifier. Second, DC adopts offline global clustering that unavoidably permutes label assignments randomly in different epochs. As a result, the classifier has to be frequently reinitialized after each label reassignment, which leads to training instability. In contrast, we optimize a non-parametric classifier at instance level and integrate the label update procedure into each training iteration with online clustering. ODC~\citep{zhan2020online} also performs online clustering. Our work differs from theirs in that ODC follows DC to optimize the cross-entropy loss between predicted and cluster labels, requiring the computationally expensive parametric classifier. In addition, ODC uses loss re-weighting to handle clustering distribution, while InterCLR directly uses online mini-batch $k$-means without resorting to other specific techniques. Experiments in Sect.~\ref{sec:exp} show that InterCLR substantially outperforms ODC.
Recently, SwAV~\citep{caron2020unsupervised} enforces consistent cluster-assignment prediction between multiple views of the same image. The cluster assignments are produced by Sinkhorn-Knopp transform~\citep{cuturi2013sinkhorn} under an equipartition constraint similar in~\cite{asano2020self}. As opposed to our work, SwAV does not compare the features but the cluster assignments of the \emph{same} instance, whereas we directly sample and compare the features of \emph{different} instances using the cluster assignments.
In conclusion, the aforementioned differences make InterCLR a simple yet effective alternative to existing clustering-based methods so that we can focus on the essence of inter-image invariance learning without the interference of other factors.

\noindent\textbf{Negative mining.}
Selection of hard negative samples has been proven effective in deep metric learning~\citep{schroff2015facenet,oh2016deep,harwood2017smart,wu2017sampling,ge2018deep,suh2019stochastic}. Some recent works~\citep{chuang2020debiased,kalantidis2020hard,robinson2021contrastive} also reveal that contrastive representation learning benefits from hard negative samples. However, prior works study the effect of negatives either in supervised learning or unsupervised intra-image invariance learning. Some works~\citep{asano2020labelling,alwassel2020self,morgado2021audio,morgado2021robust} further study the sampling issues for cross-modal invariance learning. In contrast, we target at the case of unsupervised inter-image invariance learning within a single modality, which is equally important but largely neglected. The inaccurate nature of unsupervised inter-image invariance learning makes our conclusion quite unique and complementary to existing works.

\noindent\textbf{Loss functions.}
Loss functions play an important role in unsupervised representation learning. A loss function is defined based on the properties of pretext tasks. For instance, context auto-encoders~\citep{pathak2016context} incorporate L2 loss to reconstruct input pixels, while patch orderings~\citep{doersch2015unsupervised,noroozi2016unsupervised} use cross-entropy loss to classify input image patches into pre-defined positions or orderings. Adversarial losses~\citep{goodfellow2014generative} used for representation learning are also explored in~\cite{donahue2017adversarial,donahue2019large}. The current state-of-the-art contrastive learning approaches adopt contrastive losses~\citep{hadsell2006dimensionality} that measure the similarity of sample pairs in embedding space at the instance level. Prior works~\citep{liu2017sphereface,liu2016large,wang2018cosface,wang2018additive,deng2019arcface} have shown that it is beneficial to learn discriminative features by designing large-margin loss functions in supervised learning. However, the effect of decision boundary on the unsupervised counterpart, especially for the unsupervised inter-image invariance learning scenario, is still unknown. Our work makes a first attempt \wrt decision boundary design in contrastive learning for unsupervised representations.

\section{Preliminaries}
\label{sec:prelimi}

\noindent\textbf{Intra-image invariance learning.}
A contrastive representation learning method typically learns a neural encoder $f_\theta\left(*\right)$ that maps training images $\mI=\{\vx_1,\vx_2,...,\vx_N\}$ to compact features $\mV=\{\vv_1,\vv_2,...,\vv_N\}$ with $\vv_i=f_\theta\left(\vx_i\right)$ in a $D$-dimensional L2-normalized embedding space, where the samples of a positive pair are pulled together and those of negative pairs are pushed apart.
For intra-image invariance learning, the positive pair is usually formed with two different augmented views of the same image while the negative pairs are obtained from different images.
To achieve this objective, a contrastive loss function is optimized with similarity measured by dot product.
Here we consider an effective form of contrastive loss function, called InfoNCE~\citep{oord2018representation}, as follows:
\begin{equation}\label{eq:InfoNCE}
\resizebox{.93\hsize}{!}{$
\cL_\text{InfoNCE}\!=\!\sum\limits_{i=1}^{N}-\log\frac{\exp{\left(\vv_i\cdot\vv_{i}^{+}/\tau\right)}}{\exp{\left(\vv_i\cdot\vv_{i}^{+}/\tau\right)}+\sum\limits_{\vv_{i}^{-}\in\mV_K}\!\exp{\left(\vv_i\cdot\vv_{i}^{-}/\tau\right)}},
$}
\end{equation}
where $\tau$ is a temperature hyper-parameter, $\vv_{i}^{+}$ is a positive sample for instance $i$, and $\vv_{i}^{-}\in\mV_K\subseteq\mV\setminus\{\vv_i\}$ denotes a set of $K$ negative samples randomly drawn from the training images excluding instance $i$.

\noindent\textbf{Memory bank.}
Contrastive learning requires a large number of negative samples to learn good representations~\citep{oord2018representation,wu2018unsupervised}.
However, the number of negatives is usually limited by the mini-batch size.
While simply increasing the batch to a large size (\eg, 4k-8k) can achieve good performance~\citep{chen2020simple}, it requires huge computational resources.
To address this issue, one can use a memory bank to store running average features of all samples in the dataset computed in previous steps.
Formally, let $\hat{\mV}=\{\hat{\vv}_1,\hat{\vv}_2,...,\hat{\vv}_N\}$ denote the stored features in the memory bank, these features are updated by:
\begin{equation}\label{eq:feature update}
  \hat{\vv}_i\leftarrow\left(1-\omega\right)\hat{\vv}_i+\omega\vv_i,
\end{equation}
where $\omega\in\left(0,1\right]$ is a momentum coefficient.
With a set of features $\hat{\mV}$, we can then replace $\mV$ with $\hat{\mV}$ in Eq.~\eqref{eq:InfoNCE} without having to recompute all the features every time.

\section{Methodology}
\label{sec:method}
Based on the aforementioned intra-image invariance learning, we describe how to extend the notion to leverage inter-image invariance for contrastive learning.

As shown in Fig.~\ref{fig:framework}, we introduce two invariance learning branches in our framework, one for intra-image and the other for inter-image.
The intra-image branch draws contrastive pairs by indices following the conventional protocol. The inter-image counterpart constructs contrastive pairs with pseudo-labels obtained by clustering: a positive sample for an input image is selected within the same cluster while the negative samples are obtained from other clusters.
We use variants of InfoNCE described in Sect.~\ref{sec:prelimi} as our contrastive loss and perform back-propagation to update the networks.
Within the inter-image branch, three components have non-trivial effects on learned representations and require specific designs, \ie, 1) pseudo-label maintenance, 2) sampling strategy for contrastive pairs, and 3) decision boundary design for the loss function.

\subsection{Maintaining Pseudo-Labels}
\label{subsec:pseudo-labels}

To avoid stale labels from offline clustering, we adopt mini-batch $k$-means to integrate label update into the network update iterations, thus updating the pseudo-labels \emph{on-the-fly}.

Formally, we first initialize all the features, labels and centroids via a global clustering process, \eg, $k$-means.
Next, in a mini-batch stochastic gradient descent iteration, the forward batch features are used to update the corresponding stored features in the memory bank with Eq.~\eqref{eq:feature update}.
Meanwhile, the label of each involved sample is updated by finding its current nearest centroid following:
\begin{equation}
    \min_{\vy_i\in\left\{0,1\right\}^k,\ \text{s.t.}\  \vy_i^\text{T}\vone=1}\left\|\hat{\vv}_i - \mC\vy_i\right\|_2^2,
\end{equation}
where $k$ is the number of clusters, $\mC\in\mathbb{R}^{d\times k}$ is a recorded centroid matrix with each column representing a temporary cluster centroid that evolves during training, $\vy_i$ is a $k$-dimensional one-hot vector indicating the label assignment for instance $i$.
Finally, the recorded centroid matrix is updated by averaging all the features belonging to their current and respective clusters.
In this way, labels are updated instantly along with the features.

\begin{figure}[t]
  \centering
  \includegraphics[width=.98\linewidth]{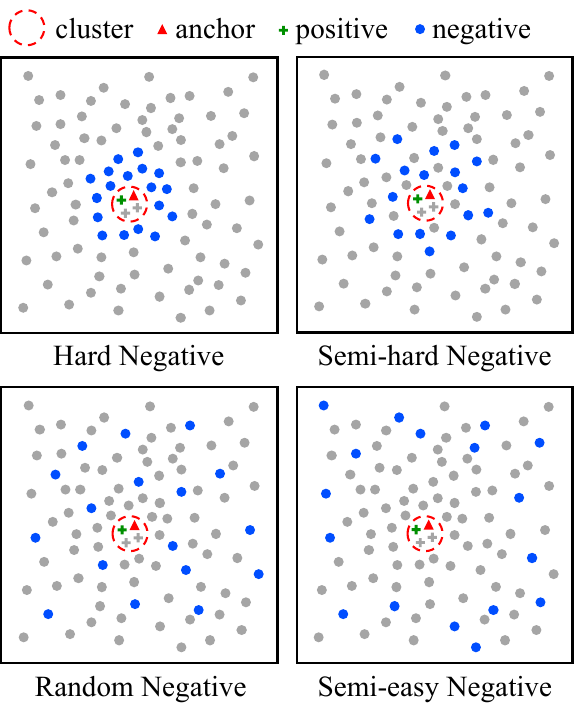}
  \caption{\textbf{Different negative sampling strategies in the embedding space.} Given an anchor (red triangle), the positive sample candidates (crosses) are those points within the cluster represented by the dashed red circle while the negative sample candidates (dots) are the points beyond this cluster. The positive sample (green cross) is drawn randomly from the cluster while the negative samples (blue dots) are drawn with different sampling strategies}
  \label{fig:sampling}
\end{figure}

\begin{figure*}[t]
  \centering
  \includegraphics[width=\linewidth]{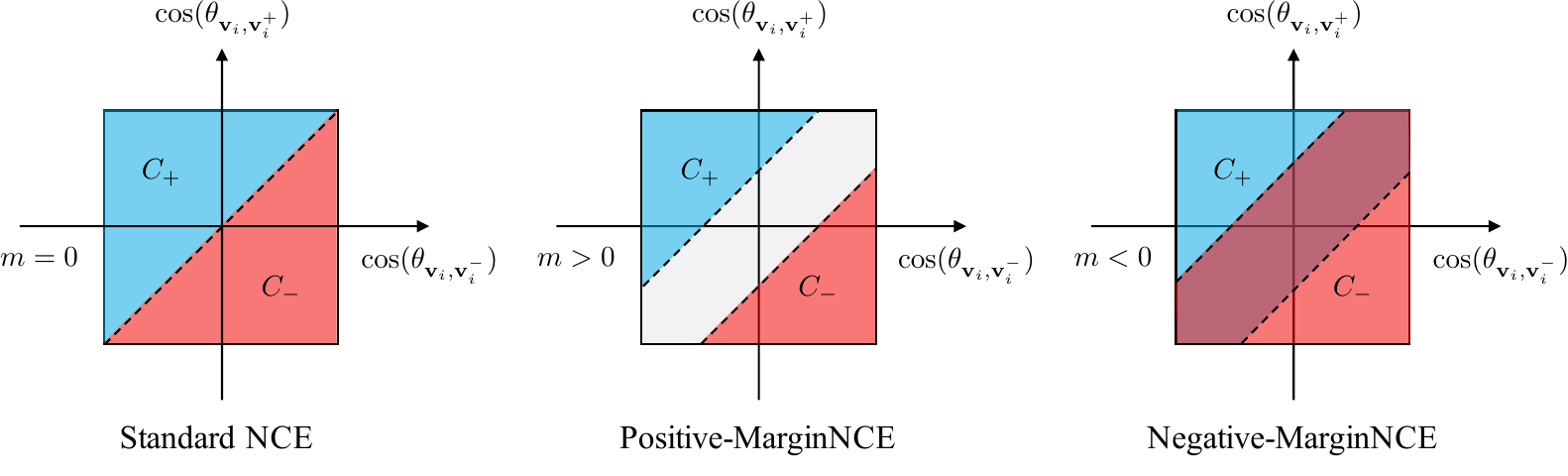}
  \caption{\textbf{Comparison of different decision margins} between standard NCE and MarginNCE under one negative sample case. The dashed line represents the decision boundary and the gray area (shown as wine red when $C_+$ and $C_-$ overlap) shows the decision margins}
  \label{fig:margin}
\end{figure*}

\subsection{Sampling Contrastive Pairs}
\label{subsec:sampling}

As discussed in Sect.~\ref{sec:intro}, sampling positive and negative pairs in unsupervised inter-image invariance learning is non-trivial. To investigate the effect of sampling, we design and compare four sampling strategies for negative samples: \emph{hard, semi-hard, random, and semi-easy}.

We define samples sharing the same label with the input image $\vx_i$ in the memory bank as positive sample candidates $\mathcal{S}_i^p$, while others as negative sample candidates $\mathcal{S}_i^n$.
For positive sampling, we randomly draw one sample from $\mathcal{S}_i^p$ and use it to form a positive pair with $\vv_i$.
For negative sampling, we sample $K$ negatives from $\mathcal{S}_i^n$.

As shown in a schematic illustration in Fig.~\ref{fig:sampling}, for ``hard negative'' sampling, we sample $K$ nearest neighbors of $\vv_i$ from $\mathcal{S}_i^n$ using cosine distance criterion.
For ``semi-hard negative'' sampling, we first create a relatively larger nearest neighbor pool, \ie, top 10\% nearest neighbors from $\mathcal{S}_i^n$, then we randomly draw $K$ samples from this pool.
For ``random negative'' sampling, we simply draw $K$ negative samples at random from $\mathcal{S}_i^n$.
For ``semi-easy negative'' sampling, similar to the ``semi-hard negative'' strategy, we first sample a pool with top 10\% farthest neighbors from $\mathcal{S}_i^n$, then we randomly draw $K$ samples from this pool\footnote{We define samples drawn within top 50\% nearest neighbors from $\mathcal{S}_i^n$ as ``semi-hard'' negatives, while top 50\% farthest neighbors from $\mathcal{S}_i^n$ as ``semi-easy'' negatives.}.
We do not include an ``easy negative'' strategy that chooses the top $K$ easiest negatives. As mentioned in Sect.~\ref{sec:intro}, the easiest samples are prone to a shortcut solution.

\subsection{Designing Decision Boundary}
\label{subsec:decision boundary}

Designing decision boundary for unsupervised inter-image invariance learning needs special care as pseudo-labels are noisy. Here, we present a way that allows decision margins to be more stringent or looser to suit the variability required in our task.

Considering the contrastive loss in Eq.~\eqref{eq:InfoNCE}, since features in the embedding space are L2-normalized, we replace $\vv_i\cdot\vv_j$ with $\cos(\theta_{\vv_i,\vv_j})$.
For simplicity of analysis, we consider the case where there is only one negative sample, \ie, a binary classification scenario.
The contrastive loss thus results in a zero-margin decision boundary given by:
\begin{equation}
    \cos(\theta_{\vv_i,\vv_{i}^{+}}) = \cos(\theta_{\vv_i,\vv_{i}^{-}}).
\end{equation}
To allow the decision margins to be more stringent or looser, we first introduce a cosine decision margin $m$ such that the decision boundary becomes:
\begin{equation}
\begin{aligned}
    C_+: \cos(\theta_{\vv_i,\vv_{i}^{+}})-m \geq \cos(\theta_{\vv_i,\vv_{i}^{-}}), \\
    C_-: \cos(\theta_{\vv_i,\vv_{i}^{-}})-m \geq \cos(\theta_{\vv_i,\vv_{i}^{+}}).
\end{aligned}
\end{equation}
As shown in Fig.~\ref{fig:margin}, $m>0$ indicates a more stringent decision boundary that encourages the discriminative ability of the representations, while $m<0$ stands for a looser decision boundary.
Then, we define a margin contrastive loss (\emph{MarginNCE}) as:
\begin{equation}
\resizebox{1.01\hsize}{!}{$
    \cL_\text{MarginNCE}\!=\!\sum\limits_{i=1}^{N}\!-\!\log\frac{\exp{\left(\left(\cos(\theta_{\vv_i,\vv_i^+})\!-\!m\right)/\tau\right)}}{\exp{\left(\left(\cos(\theta_{\vv_i,\vv_i^+})\!-\!m\right)/\tau\right)}\!+\!\sum\limits_{\vv_{i}^{-}\in\mV_K}\!\exp{\left(\cos(\theta_{\vv_i,\vv_i^-})/\tau\right)}}.
$}
\end{equation}
We make a hypothesis that for the intra-image MarginNCE loss ($\cL_\text{Intra-MarginNCE}$), the margin should be positive, since the labels derived from image indices are always correct; while for the inter-image MarginNCE loss ($\cL_\text{Inter-MarginNCE}$), the margin should be negative, since the pseudo-labels are evolving during training and are not accurate enough.
The final loss consists of these two MarginNCE loss functions:
\begin{equation}\label{eq:final loss}
\begin{aligned}
    \cL_\text{Intra-Inter-MarginNCE} = \lambda\cL_\text{Intra-MarginNCE} \\
    +\left(1-\lambda\right)\cL_\text{Inter-MarginNCE},
\end{aligned}
\end{equation}
where $\lambda$ is the weight to balance the two terms.
We study the effects of both $m$ and $\lambda$ in Sect.~\ref{subsec:empirical study}.

\section{Experiments}
\label{sec:exp}

\subsection{Implementation Details}
\label{subsec:details}

\subsubsection{Baselines}
\label{subsubsec:baselines}

Our most essential intra-image invariance learning baseline is NPID~\citep{wu2018unsupervised}: it is the special case of InterCLR by setting $\lambda=1$ in Eq.~\eqref{eq:final loss}. We find it possible to further improve the implementation of NPID by adopting more advanced techniques in~\cite{chen2020simple}: a 2-layer MLP head and stronger Gaussian blur augmentation. We denote our improved version of NPID as NPIDv2. Comparing InterCLR with NPIDv2 is critical to study the effect of inter-image invariance learning that InterCLR aims to achieve. In addition, to demonstrate that InterCLR is more generally applicable, we also experiment with most recent baselines, \ie, MoCov2~\citep{chen2020improved} and BYOL~\citep{grill2020bootstrap}.
We refer to InterCLR built upon the aforementioned baselines as NPIDv2-InterCLR, MoCov2-InterCLR and BYOL-InterCLR, respectively.

\subsubsection{Training Details}
\label{subsubsec:training}

We use ResNet-50~\citep{he2016deep} as the default backbone and perform unsupervised pre-training on the 1.28M ImageNet~\citep{deng2009imagenet} training set without labels.
Prior works~\citep{chen2020simple,grill2020bootstrap,caron2020unsupervised} have shown that using a larger batch size or training for longer epochs can improve the performance of unsupervised representations.
However, they require huge computational resources that are inaccessible to many research labs, which is not the core of this paper.
We instead focus on comparing all methods under a more commonly affordable pre-training budget, \ie, a batch size of 256 for 200 epochs with 4 NVIDIA Tesla V100 GPUs.
Besides, to demonstrate that InterCLR can also benefit from longer training epochs, we further train BYOL-InterCLR for 1000 epochs with batch size 256.
To ensure fair and direct comparisons, we generally follow \emph{the same setting} of each baseline we experiment with. The details are described next.

\noindent\textbf{NPIDv2-InterCLR.}
Our improved version of NPID~\citep{wu2018unsupervised} (\ie, NPIDv2) extends the original data augmentation in~\cite{wu2018unsupervised} by including Gaussian blur in~\cite{chen2020simple}.
However, we do not use the same heavy color distortion as~\cite{chen2020simple} since it has diminishing gains in our higher baseline.
Instead, we only apply a color jittering with a saturation factor in $[0, 2]$, and a hue factor in $[-0.5, 0.5]$.
We also add a 2-layer MLP projection head (with a 2,048-dimensional hidden layer and ReLU) to project high-dimensional features into a 128-D L2-normalized embedding space following~\cite{chen2020simple}.
We use SGD as the optimizer with a momentum of 0.9 and a weight decay of $10^{-4}$.
We adopt the cosine learning rate decay schedule~\citep{loshchilov2016sgdr} with an initial learning rate of $0.03$ using a batch size of 256 for 200 epochs.
We set the temperature parameter $\tau=0.1$, the number of negative samples $K=16,384$, and the momentum coefficient $\omega=0.5$.

The aforementioned modifications make NPIDv2 a stronger baseline, upon which we build InterCLR. Yet, NPIDv2-InterCLR substantially outperforms NPIDv2 as shown in Sect.~\ref{subsec:benchmark}.
For inter-image branch, we use online pseudo-label maintenance, semi-hard negative sampling, and cosine margin $m=-0.5$ for $\cL_\text{Inter-MarginNCE}$.
We do not use any cosine margin for intra-image branch to solely verify the effectiveness of inter-image branch.
We find over-clustering to be beneficial and set the number of clusters as 10,000, which is 10 times of the annotated number of ImageNet classes.
We set the final loss weight $\lambda=0.75$ in Eq.~\eqref{eq:final loss}.

\noindent\textbf{MoCov2-InterCLR.}
We maintain an additional memory bank to store all features from the momentum-updated encoder of MoCov2 for online clustering. We set the number of negative samples $K=16,384$, and the momentum coefficient $\omega=1$ for the memory bank. We use online pseudo-label maintenance, semi-hard negative sampling, and cosine margin $m=-0.5$ only for $\cL_\text{Inter-MarginNCE}$. We set the number of clusters as 10,000 and the final loss weight $\lambda=0.75$. Other hyper-parameters including MLP projection head, temperature parameter, training procedure and data augmentation setting exactly follow~\cite{chen2020improved}.

\begin{table*}[ht]
\centering
\small
\caption{\textbf{Image classification evaluation.} We report top-1 center-crop accuracy of fully-connected classifiers for ImageNet and Places205, and mAP of linear SVMs for VOC07 and VOC07\textsubscript{lowshot}. The parameter counts are of the feature extractors. We use officially released pre-trained model for MoCo(v2), and re-implement SimCLR and BYOL with a batch size of 256\protect\footnotemark. SimCLR, NPIDv2, MoCo(v2), BYOL and InterCLR are all pre-trained under the \emph{same} batch size and epochs for fair comparisons. Numbers with $^\dag$ are adopted from~\cite{chen2021exploring,zbontar2021barlow}. Results for SwAV are pre-trained with two 224$\times$224 views for a fair comparison. All other numbers are taken from the corresponding papers. $^\ddag$: BYOL requires a large batch size of 4096 allocated on 512 TPUs for its original reported performance. AMDIM uses FastAutoAugment~\citep{lim2019fast} that is supervised by ImageNet labels}
\label{tab:linear}
\resizebox{.97\textwidth}{!}{
\begin{tabular}{ll@{}ccccccc}
\toprule
\multirow{2}{*}{Method} & \multirow{2}{*}{Arch.} & \multirow{2}{*}{\#Params (M)} & \multirow{2}{*}{Batch size} & \multirow{2}{*}{\#Epochs} & \multicolumn{4}{c}{Transfer Dataset} \\ \cmidrule(l){6-9} 
      &     &    &    &     & ImageNet & Places205 & VOC07 &VOC07\textsubscript{lowshot}  \\ \midrule
Supervised~\citep{goyal2019scaling}  & R50 & 24 & - & - & 75.5 & 52.5 & 88.0 & 75.4 \\
Random~\citep{goyal2019scaling}  & R50 & 24 & - & - & 13.7& 16.6 & 9.6 & 9.0 \\ \hline \midrule
 \multicolumn{9}{l}{\textit{Methods using ResNet-50 within 200 training epochs:}}      \\ \midrule
Colorization~\citep{goyal2019scaling} & R50 & 24 & 640 & 28 & 39.6 & 37.5& 55.6 & 33.3 \\
Jigsaw~\citep{goyal2019scaling} & R50 & 24 & 256 & 90 & 45.7& 41.2 & 64.5 & 39.2      \\
NPID~\citep{wu2018unsupervised} & R50 & 24 & 256 & 200 & 54.0 & 45.5 & - & - \\
Rotation~\citep{gidaris2018unsupervised} & R50 & 24 & - & - & 48.9& 41.4 & 63.9 & - \\
BigBiGAN~\citep{donahue2019large} & R50 & 24 & - & - & 56.6 & - & - & -\\
LA~\citep{zhuang2019local} & R50 & 24 & 128 & 200 & 58.8 & 49.1& - & - \\
CPC v2~\citep{henaff2019data} & R50 & 24 & 512 & 200 & 63.8 & - & -&- \\
MoCo~\citep{he2020momentum} & R50 & 24 & 256 & 200 & 60.6 & 50.2 & 79.3 & 57.9 \\
SimCLR~\citep{chen2020simple} & R50 & 24 & 256 & 200 & 61.9 & 51.6 & 79.0 & 58.4 \\
PCL v2~\citep{li2021prototypical} & R50 & 24 & 256 & 200 & 67.6 & 50.3 & 85.4 & - \\
SwAV~\citep{caron2020unsupervised} & R50 & 24 & 4096 & 200 & \; 69.1$^\dag$ & - & - &- \\
SimSiam~\citep{chen2021exploring} & R50 & 24 & 256 & 200 & 70.0 & - & - & - \\
InfoMin Aug.~\citep{tian2020makes} & R50 & 24 & 256 & 200 & 70.1 & - & - &- \\
NNCLR~\citep{dwibedi2021little} & R50 & 24 & 4096 & 200 & 70.7 & - & - &- \\ \midrule
NPIDv2  & R50  & 24  & 256 & 200 & 64.6  & 51.9  & 81.7  & 63.2 \\
\rowcolor{tabhighlight} NPIDv2-InterCLR  & R50 & 24 & 256 & 200 & 65.7 & 52.4 & 82.8 & 65.8  \\ \midrule
MoCov2~\citep{chen2020improved} & R50 & 24 & 256 & 200 & 67.5 & 52.5 & 84.2 & 68.2 \\
\rowcolor{tabhighlight} MoCov2-InterCLR  & R50 & 24 & 256 & 200 & 68.0 & 52.6 & 85.3 & \textbf{70.7}  \\ \midrule
BYOL~\citep{grill2020bootstrap} & R50 & 24 & 256 & 200 & 70.6 & 52.7 & 85.1 & 68.9 \\
\rowcolor{tabhighlight} BYOL-InterCLR & R50 & 24 & 256 & 200 & \textbf{73.5} & \textbf{53.8} & \textbf{86.5} & 69.6 \\ \hline \midrule
 \multicolumn{9}{l}{\textit{Methods using other architectures or longer training epochs:}}      \\ \midrule
SeLa~\citep{asano2020self} & R50 & 24 & 256 & 400 & 61.5 & - & - &-\\
ODC~\citep{zhan2020online} & R50 & 24 & 512 & 440 & 57.6 & 49.3 & 78.2 & 57.1 \\
PIRL~\citep{misra2020self} & R50 & 24 & 1024 & 800 & 63.6  & 49.8 & 81.1 &-\\
SimCLR~\citep{chen2020simple} & R50 & 24 & 4096 & 1000 & 69.3 & \; 52.5$^\dag$ & \; 85.5$^\dag$ &-\\
SwAV~\citep{caron2020unsupervised} & R50 & 24 & 4096 & 800 & \; 71.8$^\dag$ & \; 52.8$^\dag$ & \; 86.4$^\dag$ &- \\
SimSiam~\citep{chen2021exploring} & R50 & 24 & 256 & 800 & 71.3 & - & - &- \\
InfoMin Aug.~\citep{tian2020makes} & R50 & 24 & 256 & 800 & 73.0 & - & - &- \\
NNCLR~\citep{dwibedi2021little} & R50 & 24 & 256 & 1000 & 68.7 & - & - &- \\
Barlow Twins~\citep{zbontar2021barlow} & R50 & 24 & 2048 & 1000 & 73.2 & 54.1 & 86.2 &- \\
BYOL$^\ddag$~\citep{grill2020bootstrap} & R50 & 24 & 4096 & 1000 & 74.3 & \; 54.0$^\dag$ & \; 86.6$^\dag$ &- \\ \midrule
NPIDv2  & R50  & 24  & 256 & 1000 & 68.2  & 52.8  & 84.6  & 68.3 \\
\rowcolor{tabhighlight} NPIDv2-InterCLR  & R50 & 24 & 256 & 1000 & 69.6 & 53.4 & 85.7 & \textbf{70.0}  \\ \midrule
BYOL~\citep{grill2020bootstrap} & R50 & 24 & 256 & 1000 & 73.4 & 53.6 & 86.1 & 69.0 \\ 
\rowcolor{tabhighlight} BYOL-InterCLR & R50 & 24 & 256 & 1000 & \textbf{74.5} & \textbf{54.4} & \textbf{87.0} & \textbf{70.0} \\ \midrule
CPC~\citep{oord2018representation} & R101 & 28 & 512 & - & 48.7 & - & - &-\\
CMC~\citep{tian2020contrastive} & R50$_\text{L+ab}$ & 47 & 128 & 240 & 64.0 & - & - &-\\
AMDIM$^\ddag$~\citep{bachman2019learning} & Custom & 626 & 1008 & 150 & 68.1 & 55.0 & -  & - \\ \bottomrule
\end{tabular}
}
\end{table*}

\noindent\textbf{BYOL-InterCLR.}
We exactly follow~\cite{grill2020bootstrap} for the training hyper-parameters and augmentation recipes.
Considering that many previous methods report their performance on 200 epochs, we also train for 200 epochs for fair comparisons, following the 300-epoch recipes in~\cite{grill2020bootstrap}: base learning rate is 0.3, weight decay is $10^{-6}$, and the base target exponential moving average parameter is 0.99. The same 200-epoch recipes are also adopted in~\cite{chen2021exploring}.
For 1000-epoch pre-training, we follow the same 1000-epoch recipes in~\cite{grill2020bootstrap}: base learning rate is 0.2, weight decay is $1.5\times10^{-6}$, and the base target exponential moving average parameter is 0.996.
Similar to MoCov2-InterCLR, we use a memory bank to store all features from the target network of BYOL to facilitate online clustering.
\footnotetext{We accumulate gradients to simulate batch size 4096 for BYOL experiments due to resource constraints. Thus, our reproduced 1000-epoch results are relatively lower than the original reported performance. Nevertheless, all experiments are done under the same setting for fair comparisons.}
Since BYOL uses a mean squared error loss without negative samples, we do not use any negative sampling and cosine margin in this entry. Instead, we only sample the positive pair after clustering and adopt the same loss in place of the MarginNCE loss for inter-image invariance learning. Following NPIDv2 and MoCov2 experiments, we set the number of clusters as 10,000 and the final loss weight $\lambda=0.75$. Although no negative pairs are sampled for inter-image branch, we empirically observe no collapsing solutions due to the stop-gradient operation introduced in BYOL.

\subsection{Results on Standard Benchmarks}
\label{subsec:benchmark}

Following common practice in unsupervised representation learning~\citep{zhang2017split,goyal2019scaling}, we evaluate the quality of learned representations by transferring them to multiple downstream tasks.
We perform experiments on a variety of datasets, focusing on four kinds of downstream tasks:
image classification with linear models (Sect.~\ref{subsubsec:linear}), low-shot image classification (Sect.~\ref{subsubsec:low-shot}), semi-supervised learning (Sect.~\ref{subsubsec:semi-sup}), and object detection (Sect.~\ref{subsubsec:detection}).
Our evaluations cover two learning setups where the pre-trained network is either \emph{frozen} as a feature extractor (Sect.~\ref{subsubsec:linear} and Sect.~\ref{subsubsec:low-shot}), or fully \emph{fine-tuned} as weight initialization (Sect.~\ref{subsubsec:semi-sup} and Sect.~\ref{subsubsec:detection}).

\subsubsection{Image Classification with Linear Models}
\label{subsubsec:linear}

\noindent\textbf{Setup.}
Following~\cite{goyal2019scaling,misra2020self}, we freeze all the backbone parameters and train classifiers on representations from different depths of the network on three datasets, including ImageNet~\citep{deng2009imagenet}, Places205~\citep{zhou2014learning} and VOC07~\citep{everingham2010pascal}.
For ImageNet and Places205, we train a 1000-way and 205-way fully-connected classifier, respectively, on the frozen representations using SGD with a momentum of 0.9.
For ImageNet, we train for 100 epochs, with a batch size of 256 and a weight decay of $10^{-4}$.
The learning rate is initialized as 0.01, decayed by a factor of 10 after every 30 epochs.
For Places205, we train for 14 epochs, with a batch size of 256 and a weight decay of $10^{-4}$.
The learning rate is initialized as 0.01, dropped by a factor of 10 at three equally spaced intervals.
We report top-1 center-crop accuracy on the official validation split of ImageNet and Places205.
For VOC07, we follow the same setup in~\cite{goyal2019scaling,misra2020self} and train linear SVMs on the frozen representations using LIBLINEAR package~\citep{fan2008liblinear}.
We train on the trainval split of VOC07 and report mAP on the test split.

\noindent\textbf{Results.}
Table~\ref{tab:linear} shows the results for the best-performing layer of each method.
Our improved NPIDv2 baseline already performs well on the three datasets.
However, InterCLR substantially outperforms NPIDv2, demonstrating the benefits of introducing inter-image invariance.
Similar improvements are also observed in MoCov2 and BYOL.
By building upon a stronger baseline in intra-image invariance learning, InterCLR outperforms previous unsupervised learners that are pre-trained within 200 epochs using a feasible 256 batch size on a standard ResNet-50 backbone, setting a new state of the art in this fair competition on all three datasets.
Note that our 200-epoch BYOL-InterCLR has already yielded better performance than 1000-epoch BYOL, indicating that the performance efficiency of InterCLR is at least 5$\times$ than BYOL below 1000 epochs.
When pre-trained for 1000 epochs, InterCLR still consistently outperforms its corresponding intra-image invariance learning baseline by clear margins.
Both the higher performance efficiency and the final performance demonstrate that InterCLR helps learn generalizable representations \emph{faster} and \emph{better}.

\begin{figure}[t]
    \centering
	\includegraphics[width=\linewidth]{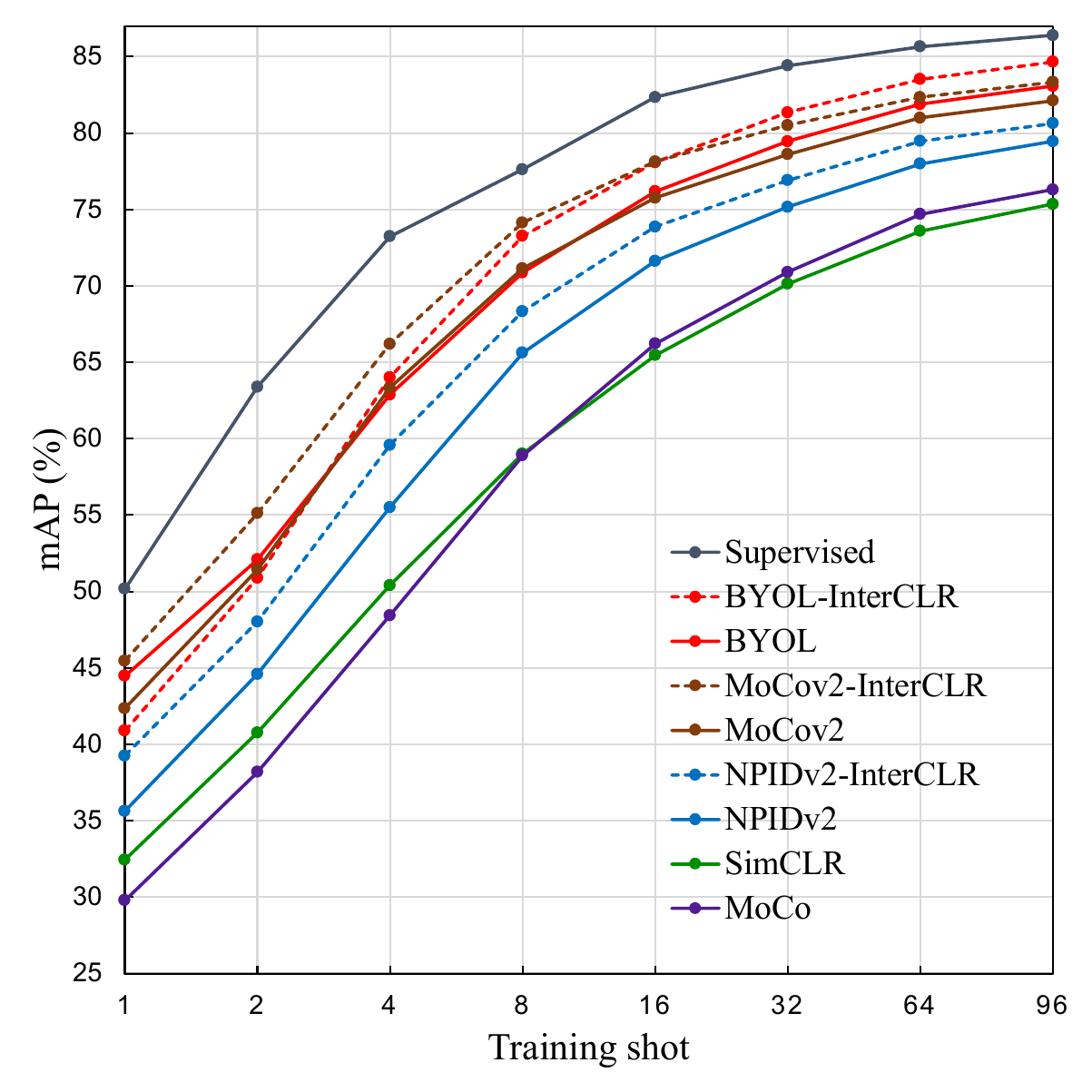}
	\caption{\textbf{Low-shot image classification} on VOC07 with linear SVMs trained on the features from the best-performing layer of each method for ResNet-50. All unsupervised methods are pre-trained for 200 epochs on ImageNet for fair comparisons. We show the average performance for each shot across five runs. Results for MoCo(v2) are evaluated using the officially released pre-trained model. Results for SimCLR and BYOL are re-implemented by us}
	\label{fig:lowshot}
\end{figure}

\subsubsection{Low-Shot Image Classification}
\label{subsubsec:low-shot}

\noindent\textbf{Setup.}
Next, we explore the quality of learned representations when there are few training examples per category by transferring to the low-shot VOC07 classification task.
Specifically, we vary the number of labeled examples in each class and train linear SVMs on the frozen backbone following the same procedure in~\cite{goyal2019scaling}.
We train on the trainval split of VOC07 and report mAP across five independent samples for each low-shot value evaluated on the test split of VOC07.

\noindent\textbf{Results.}
Table~\ref{tab:linear} shows the final mAP results of different methods obtained with the averages of all low-shot values and all independent runs.
InterCLR substantially outperforms its intra-image invariance learning counterpart.
Fig.~\ref{fig:lowshot} also provides the per-shot results pre-trained within 200 epochs.
InterCLR improves upon all baselines and gradually bridges the gap to supervised pre-training when the number of labeled examples per class is increasing.

\begin{table}[t]
\centering
\small
\caption{\textbf{Semi-supervised learning} on ImageNet. We report top-5 center-crop accuracy on the ImageNet validation set of unsupervised models that are fine-tuned with 1\% and 10\% of the labeled ImageNet training data. We use officially released pre-trained model for MoCo(v2), and re-implement SimCLR and BYOL. All other numbers are taken from the corresponding papers}
\addtolength{\tabcolsep}{-2pt}
\resizebox{.48\textwidth}{!}{
\begin{tabular}{llccc}
\toprule
\multirow{3}{*}{Method} & \multirow{3}{*}{Backbone} & \multirow{3}{*}{\#Epochs} & \multicolumn{2}{c}{Label fraction} \\
               &         &       & 1\%                & 10\%              \\
               &         &       & \multicolumn{2}{c}{Top-5 accuracy}     \\ \midrule
Supervised~\citep{zhai2019s4l}  & R50  & -  & 48.4  & 80.4 \\
Random~\citep{wu2018unsupervised} & R50 & - & 22.0  & 59.0 \\ \hline \midrule
\multicolumn{5}{l}{\textit{Methods using semi-supervised learning:}}      \\ \midrule
Pseudo-label~\citep{lee2013pseudo}   & R50v2   & -     & 51.6               & 82.4              \\
VAT + Ent Min.~\citep{miyato2018virtual} & R50v2        & -      & 47.0                   & 83.4 \\ 
S$^4$L Exemplar~\citep{zhai2019s4l} & R50v2        & -      & 47.0                   & 83.7 \\
S$^4$L Rotation~\citep{zhai2019s4l} & R50v2        & -      & 53.4                  & 83.8 \\ \hline \midrule
\multicolumn{5}{l}{\textit{Methods using unsupervised learning only:}} \\ \midrule
NPID~\citep{wu2018unsupervised}           & R50     & 200   & 39.2               & 77.4              \\
Jigsaw~\citep{goyal2019scaling}        & R50     & 90   & 45.3               & 79.3              \\
MoCo~\citep{he2020momentum}        & R50     & 200   & 61.3        & 84.0             \\
SimCLR~\citep{chen2020simple}  & R50  & 200   & 64.5        & 82.6              \\
PCL v2~\citep{li2021prototypical}  & R50  & 200   & 73.9        & 85.0             \\ \midrule
NPIDv2  & R50  & 200  & 63.0  & 84.0 \\
\rowcolor{tabhighlight} NPIDv2-InterCLR         & R50     & 200     & 65.8             & 84.5                 \\ \midrule
MoCov2~\citep{chen2020improved} & R50  & 200  & 67.7 & 85.0 \\
\rowcolor{tabhighlight} MoCov2-InterCLR         & R50  & 200 & 72.7 & 85.9  \\ \midrule
BYOL~\citep{grill2020bootstrap} & R50  & 200  & 76.8  & 87.8  \\
\rowcolor{tabhighlight} BYOL-InterCLR          & R50  & 200 & \textbf{79.4} & \textbf{89.2}  \\ \hline \midrule
PIRL~\citep{misra2020self}         & R50     & 800   & 57.2               & 83.8             \\
SimCLR~\citep{chen2020simple}       & R50     & 1000   & 75.5               & 87.8              \\
SwAV~\citep{caron2020unsupervised}       & R50     & 800   & 78.5               & 89.9              \\
Barlow Twins~\citep{zbontar2021barlow}  & R50     & 1000   & 79.2               & 89.3              \\
NNCLR~\citep{dwibedi2021little}        & R50     & 1000   & \textbf{80.7}               & 89.3              \\ \midrule
NPIDv2  & R50  & 1000  & 77.2  & 88.1 \\
\rowcolor{tabhighlight} NPIDv2-InterCLR         & R50     & 1000     & 78.6             & 88.8                 \\ \midrule
BYOL~\citep{grill2020bootstrap}       & R50     & 1000   & 78.4               & 89.0              \\
\rowcolor{tabhighlight} BYOL-InterCLR    & R50      & 1000   & 80.5 & \textbf{90.2}  \\ \bottomrule
\end{tabular}
}
\label{tab:semi-sup}
\end{table}

\begin{table}[ht]
\centering
\small
\caption{\textbf{Object detection} fine-tuned on VOC07+12 using Faster-RCNN. We report AP$_{50}$, the default metric for VOC object detection, averaged over five trials. All unsupervised methods are pre-trained for 200 epochs on ImageNet for fair comparisons. We also append the results of some methods pre-trained for longer epochs as a reference. Most numbers are taken from~\cite{he2020momentum,chen2021exploring}}
\addtolength{\tabcolsep}{-2pt}
\resizebox{.48\textwidth}{!}{
\begin{tabular}{llcc}
\toprule
Method     & Architecture  & \#Epochs & VOC07+12  \\ \midrule
Random~\citep{he2020momentum} &  R50-C4 & - & 60.2 \\
Supervised~\citep{he2020momentum} & R50-C4    & 90         & 81.3    \\ \midrule
NPID~\citep{wu2018unsupervised}   & R50-C4    & 200         & 80.9    \\ 
PIRL~\citep{misra2020self}   & R50-C4    & 200         & 81.0    \\ 
MoCo~\citep{he2020momentum}       & R50-C4    & 200         & 81.5        \\
SimCLR~\citep{chen2020simple}     & R50-C4    & 200          & 81.8       \\
SwAV~\citep{caron2020unsupervised}   & R50-C4    & 200         & 81.5       \\
SimSiam~\citep{chen2021exploring}  & R50-C4    & 200         & 82.4 \\ \midrule
NPIDv2                           & R50-C4   & 200          & 81.5       \\
\rowcolor{tabhighlight} NPIDv2-InterCLR   & R50-C4      & 200       & 81.9       \\ \midrule
MoCov2~\citep{chen2020improved}   & R50-C4   & 200          & 82.4       \\
\rowcolor{tabhighlight} MoCov2-InterCLR   & R50-C4      & 200       & \textbf{82.7}       \\ \midrule
BYOL~\citep{grill2020bootstrap}           & R50-C4     & 200        & 81.4       \\
\rowcolor{tabhighlight} BYOL-InterCLR   & R50-C4    & 200         & \textbf{82.7}       \\ \midrule
MoCov2~\citep{chen2020improved}   & R50-C4   & 800          & 82.5       \\
SwAV~\citep{caron2020unsupervised}   & R50-C4    & 800         & 82.6       \\ 
Barlow Twins~\citep{zbontar2021barlow}  & R50-C4    & 1000         & 82.6       \\ \bottomrule
\end{tabular}
}
\label{tab:detection}
\end{table}

\subsubsection{Semi-Supervised Learning}
\label{subsubsec:semi-sup}

\noindent\textbf{Setup.}
We perform semi-supervised learning on ImageNet following~\cite{wu2018unsupervised,misra2020self}.
We randomly select 1\% and 10\% subsets of the labeled ImageNet training data in a class-balanced way.
Then, we fine-tune our models on these two subsets.
More specifically, we use the 1\% and 10\% ImageNet subsets specified in the official code release of SimCLR~\citep{chen2020simple}.
For both 1\% and 10\% labeled data, we fine-tune using SGD with a momentum of 0.9 and a batch size of 256 for 20 epochs, with the initial learning rate of backbone set as 0.01 and that of linear classifier as 1.
The learning rate is decayed by a factor of 5 at 12 and 16 epochs.
We use a weight decay of $5\times10^{-4}$ for 1\% fine-tuning and $10^{-4}$ for 10\% fine-tuning.
We report top-5 center-crop accuracy on the official ImageNet validation split.

\noindent\textbf{Results.}
As shown in Table~\ref{tab:semi-sup}, InterCLR again boosts the performance of all tested intra-image invariance learners by clear margins, especially when only 1\% labeled data is available.
We also observe that BYOL-InterCLR trained for only 200 epochs can achieve better results than BYOL trained for 1000 epochs.
The 200-epoch results are even comparable with prior arts trained with a much larger compute.
With 1000-epoch pre-training, InterCLR further improves the performance, performing on par (1\% labels) or even better (10\% labels) than state-of-the-art results in this more challenging low-data regime.

\subsubsection{Object Detection}
\label{subsubsec:detection}

\noindent\textbf{Setup.}
Following~\cite{he2020momentum}, we use Detectron2~\citep{wu2019detectron2} to train the Faster-RCNN~\citep{ren2015faster} object detection model with a R50-C4 backbone~\citep{he2017mask}, with BatchNorm tuned.
Specifically, we use a batch size of 2 images per GPU, a total of 8 GPUs and fine-tune ResNet-50 models for 24k iterations ($\sim$23 epochs).
The learning rate is initialized as 0.02 with a linear warmup for 100 iterations, and decayed by a factor of 10 at 18k and 22k iterations.
The image scale is $[480, 800]$ pixels during training and 800 at inference.
Following~\cite{chen2021exploring}, we also search the fine-tuning learning rate for BYOL experiments.
We fine-tune all layers on the trainval split of VOC07+12, and evaluate on the test split of VOC07.
We use the same setup for both supervised and unsupervised models.

\noindent\textbf{Results.}
The results averaged across five runs are summarized in Table~\ref{tab:detection}.
InterCLR consistently outperforms all examined intra-image invariance learning baselines pre-trained for 200 epochs, with the largest improvement observed in BYOL (+1.3\%).
InterCLR also surpasses the supervised pre-training by 1.4\%.
It should be noted that our 200-epoch results (\ie, MoCov2-InterCLR and BYOL-InterCLR) are even better than other results obtained with much longer pre-training epochs.

\subsection{Empirical Study}
\label{subsec:empirical study}

\begin{figure}[t]
  \centering
  \includegraphics[width=\linewidth]{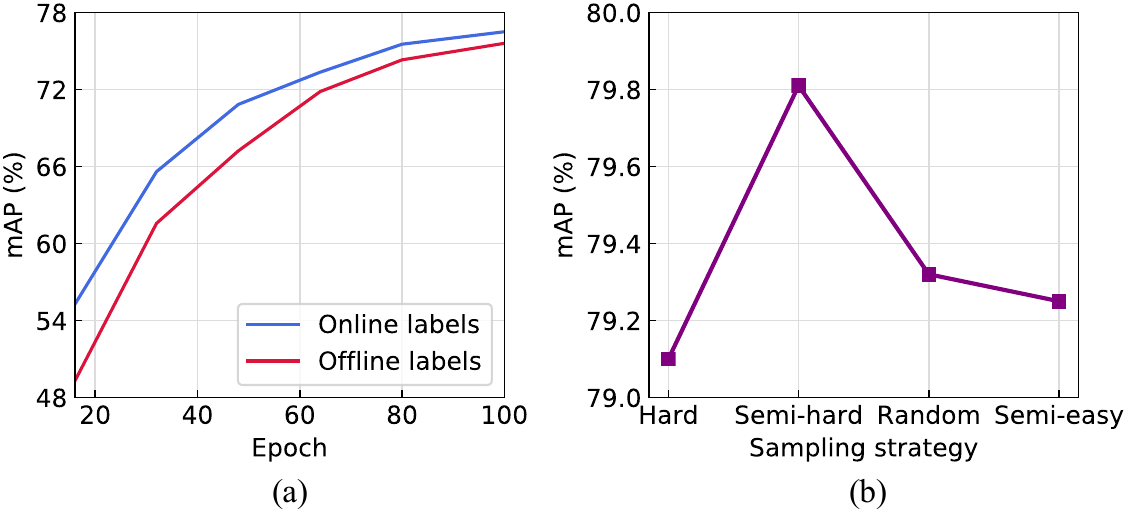}
  \caption{\textbf{(a)} Comparison between online labels and offline labels. \textbf{(b)} Comparison of different sampling strategies. We report mAP of linear SVMs on the VOC07 classification benchmark}
  \label{fig:expstudy_label_sampling}
\end{figure}

We conduct a comprehensive empirical study for inter-image invariance learning in this subsection.
To perform a large amount of experiments needed for the study, we adjust the experimental setting to train each model with fewer (4,096) negative samples or for fewer (100) epochs while keeping the other hyper-parameters in Sect.~\ref{subsec:details} unchanged\footnote{In practice, we observe that adopting 4096 negative samples or 100 pre-training epochs does not affect our empirical observations.}.

Specifically, when studying the three main aspects (\ie, pseudo-label maintenance, sampling strategy, and decision boundary design) in Sect.~\ref{subsubsec:properties}, we train with 4,096 negative samples for 100 epochs and perform a set of experiments progressively:
for pseudo-label maintenance study, we use random negative sampling and zero-margin decision boundary;
for sampling strategy study, we use online pseudo-label maintenance and zero-margin decision boundary;
for decision boundary study, we use online pseudo-label maintenance and semi-hard negative sampling.
When conducting the further analysis in Sect.~\ref{subsubsec:further analysis}, we adopt exactly the same setting as in Sect.~\ref{subsec:details} (\ie, 200-epoch pre-training with 16,384 negative samples), except for the ablation on loss weight $\lambda$ in Eq.~\eqref{eq:final loss}, where we train with 4,096 negative samples and use online pseudo-label maintenance, random negative sampling as well as zero-margin decision boundary.
As a result, we obtain relatively lower performances in most cases.
However, these experiments aim at better understanding the properties of InterCLR and provide a useful guidance on how to design each component for inter-image invariance learning.
Throughout the section, we take NPIDv2-InterCLR as the prototype for empirical study and use the standard benchmarks from Sect.~\ref{subsec:benchmark} to measure the quality of learned representations.

\begin{figure}[t]
  \centering
  \includegraphics[width=.96\linewidth]{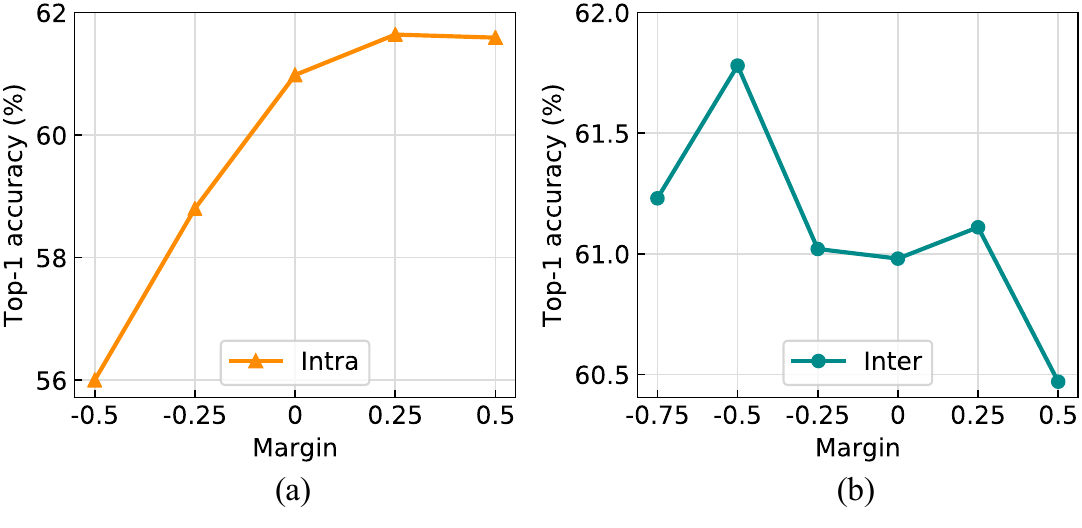}
  \caption{Effect of decision margin for \textbf{(a)} intra-image branch, and \textbf{(b)} inter-image branch. We report top-1 accuracy on the ImageNet linear classification benchmark}
  \label{fig:expstudy_margin}
\end{figure}

\subsubsection{Main Observations}
\label{subsubsec:properties}

\noindent\textbf{Observation 1: Online labels converge faster and perform better than offline labels.}
We compare the effect of two investigated pseudo-label maintenance mechanisms (\ie, online mini-batch $k$-means \textit{vs.} offline global $k$-means) on the learned representations.
As shown in Fig.~\ref{fig:expstudy_label_sampling}(a), we observe that \emph{online labels achieve faster convergence and better performance than offline labels during the training process}.
Due to the expensive computational cost, the sparsely updated pseudo-labels of offline global $k$-means are rather stale relative to the rapidly updated network. Thus, offline labels are less reliable than instantly updated online labels as the latter is simultaneously undertaken along with the network update.
This suggests the superiority of maintaining pseudo-labels online for inter-image invariance learning.

\noindent\textbf{Observation 2: Semi-hard negative sampling is more reliable and unbiased.}
We then study the importance of negative sampling strategies.
Fig.~\ref{fig:expstudy_label_sampling}(b) compares four negative sampling strategies discussed in Sect.~\ref{subsec:sampling}.
Interestingly, we find that semi-hard negative sampling achieves the best performance, while hard negative sampling is even worse than the na\"ive random sampling strategy.
In the context of unsupervised inter-image invariance learning, the hard negatives are likely to have false labels due to the noisy cluster assignments, \ie, they may actually be positive pairs. Solely mining hard negatives will intensify the bias during training. In contrast, semi-hard negative sampling provides the chance to correct these false negatives while maintaining the hardness of sampled negatives at the same time.
In conclusion, the observation reveals that \emph{different from what is commonly adopted in supervised learning, hard negative mining is not the best choice in the unsupervised scenario; on the contrary, randomly sampled negatives within a relatively larger nearest neighbor pool are more reliable and unbiased for unsupervised inter-image invariance learning}.

\begin{figure}[t]
  \centering
  \includegraphics[width=.9\linewidth]{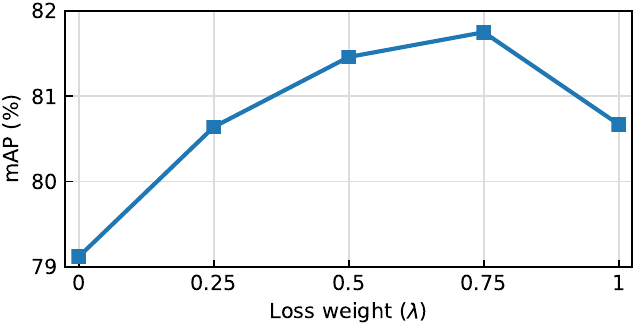}
  \caption{\textbf{Effect of loss weight $\lambda$ in Eq.~\eqref{eq:final loss}} on the quality of learned representations. We report mAP of linear SVMs on the VOC07 classification benchmark}
  \label{fig:lossweight}
\end{figure}

\noindent\textbf{Observation 3: Positive decision margin for ``Intra'' and negative decision margin for ``Inter''.}
We study the impact of decision boundary using different cosine margins.
Specifically, we perform a set of margin experiments for each branch by setting the margin of the other branch as $0$.
For the intra-image branch shown in Fig.~\ref{fig:expstudy_margin}(a), using a positive margin improves the performance, while a negative one degrades the performance.
Since the labels derived from image indices are always correct, it is beneficial to adopt a positive margin to further strengthen the decision boundary.
Hence, it is necessary to pursue highly discriminative features for intra-image invariance learning, which is in accordance with the case of supervised learning.
However, the opposite phenomenon is observed for the inter-image branch as shown in Fig.~\ref{fig:expstudy_margin}(b).
Using a negative margin (the best performance is observed when $m=-0.5$) improves upon zero margin, while a positive one fluctuates and even degrades the performance.
We attribute this phenomenon to the inaccurate nature of inter-image invariance learning. The pseudo-labels derived from the inter-image branch are evolving during training and are noisy at initial epochs, \ie, they usually contain false positive/negative samples. Adopting a positive margin will intensify the faulty cases and then the situation will never be corrected. In contrast, a negative margin can mitigate the inaccuracy during training, leading to a more stable evolution of pseudo-labels.
Therefore, \emph{rather than solely pursue discriminative features, it is beneficial to design a less stringent decision boundary for inter-image invariance learning}.

\subsubsection{Further Analysis}
\label{subsubsec:further analysis}

\noindent\textbf{Effect of loss weight $\lambda$.}
The benefits of inter-image invariance brought by InterCLR have been fully demonstrated on various downstream tasks in Sect.~\ref{subsec:benchmark}.
Here, we further analyze the effect of $\lambda$ that controls the weight between two MarginNCE losses in Eq.~\eqref{eq:final loss}.
For $\lambda=1$, our framework degenerates to a typical form of intra-image invariance learning in~\cite{wu2018unsupervised}.
For $\lambda=0$, only the inter-image invariance learning branch is retained.
InterCLR benefits from the combination of two kinds of image invariance learning as shown in Fig.~\ref{fig:lossweight}, with the best trade-off obtained when $\lambda=0.75$.

\begin{figure}[t]
  \centering
  \includegraphics[width=\linewidth]{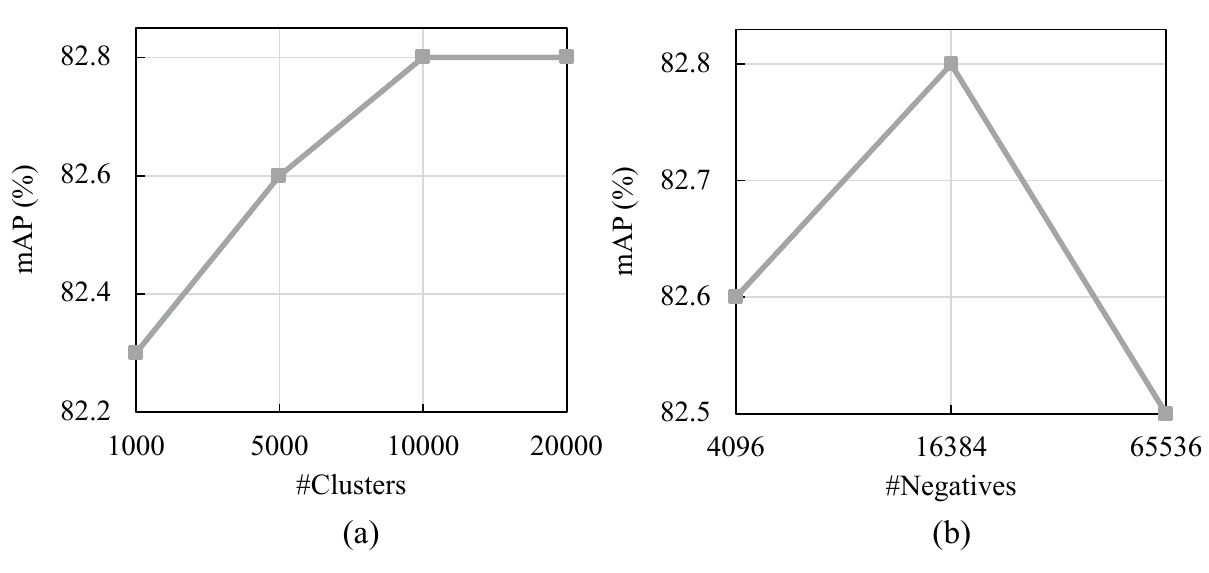}
  \caption{Effect of the number of \textbf{(a)} clusters, and \textbf{(b)} negative samples on the quality of learned representations. We report mAP of linear SVMs on the VOC07 classification benchmark}
  \label{fig:expstudy_cluster_negative}
\end{figure}

\begin{table}[t]
\centering
\small
\caption{\textbf{Computational cost comparison} on 8 V100 GPUs. The training cost is measured based on 200 epochs}
\addtolength{\tabcolsep}{-2pt}
\label{tab:cost}
\begin{tabular}{lccc}
\toprule
Method          & Batch size & Memory / GPU & Time / 200-ep. \\ \midrule
Intra          & 256           & 5.6 G           & 2.6 days             \\
Intra+Inter (offline) & 256           & 6.1 G           & 6.4 days             \\
\rowcolor{tabhighlight} Intra+Inter (online)  & 256           & 6.2 G           & 5 days             \\ \bottomrule
\end{tabular}
\end{table}

\begin{figure*}[t]
    \centering
	\includegraphics[width=\linewidth]{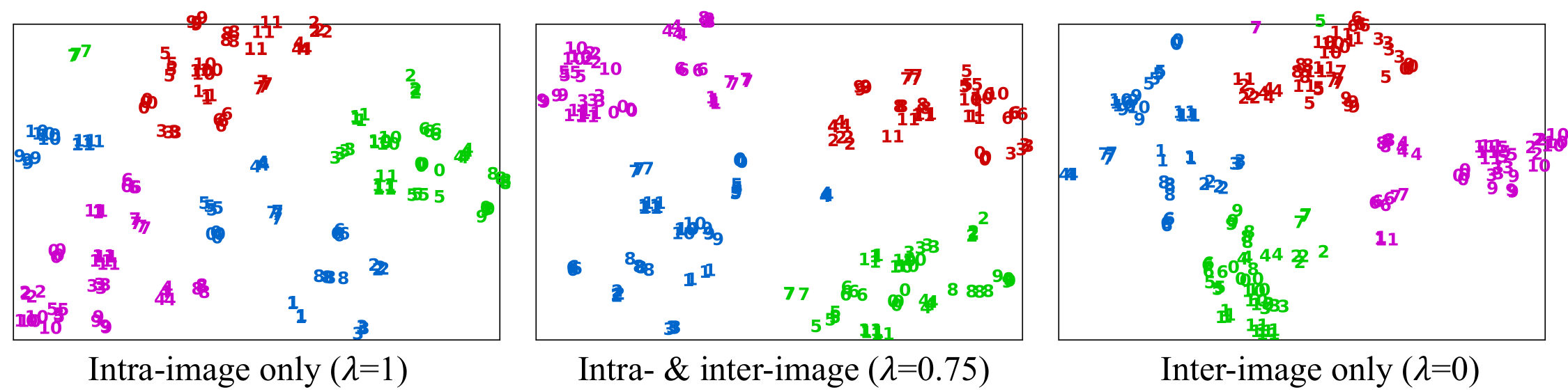}
	\caption{\textbf{Feature space visualization via t-SNE.} Colors indicate ImageNet original classes, and numbers indicate different images in each class. Points with the same color and number are the same image in different augmentations}
	\label{fig:tsne}
\end{figure*}

\noindent\textbf{Effect of the number of clusters.}
We study the effect of the number of clusters on the quality of learned representations. The results are presented in Fig.~\ref{fig:expstudy_cluster_negative}(a). Over-clustering tends to be beneficial for inter-image invariance learning while the performance gradually saturates around 10,000, which is 10 times of the annotated number of original ImageNet classes.

\noindent\textbf{Effect of negative samples.}
Prior works~\citep{oord2018representation,wu2018unsupervised,he2020momentum,chen2020simple} have shown that intra-image invariance learning benefits greatly from a larger number of negative samples. We examine whether this trend still holds for inter-image invariance learning. As shown in Fig.~\ref{fig:expstudy_cluster_negative}(b), increasing the number of negatives from 4,096 to 16,384 only has marginal benefits on the final performance. Using a larger number of negative samples (\eg, 65,536) even degrades the performance. This demonstrates that inter-image invariance learning is more robust to the number of negative samples, \ie, we can mitigate the reliance of large negative samples for intra-image invariance learning by incorporating the inter-image branch.

\noindent\textbf{Computational cost.}
In Table~\ref{tab:cost}, we compare the memory and time cost of intra- and inter-image invariance learning. Compared with the intra-image invariance learning baseline, \ie, Intra, incorporating inter-image invariance learning inevitably increases the training cost. Nevertheless, maintaining iteration-based online labels, \ie, Intra+Inter (online), is around $\times 1.3$ faster than commonly adopted epoch-based offline labels, \ie, Intra+Inter (offline).

\noindent\textbf{Feature space visualization.}
Apart from quantitative results, we also visualize the learned representations in the t-SNE~\citep{maaten2008visualizing} embedding space.
As shown in Fig.~\ref{fig:tsne}, the ``Intra-image only'' model merely groups the same image in different augmentations together; however, different images are separated even though they are in the same class.
The ``Inter-image only'' model shortens the distance between images in the same class; however, outliers emerge.
The ``Intra- \& inter-image'' model well inherits the advantages from above two models, resulting in a more separable feature space.

\begin{figure*}[ht]
  \centering
  \includegraphics[width=\linewidth]{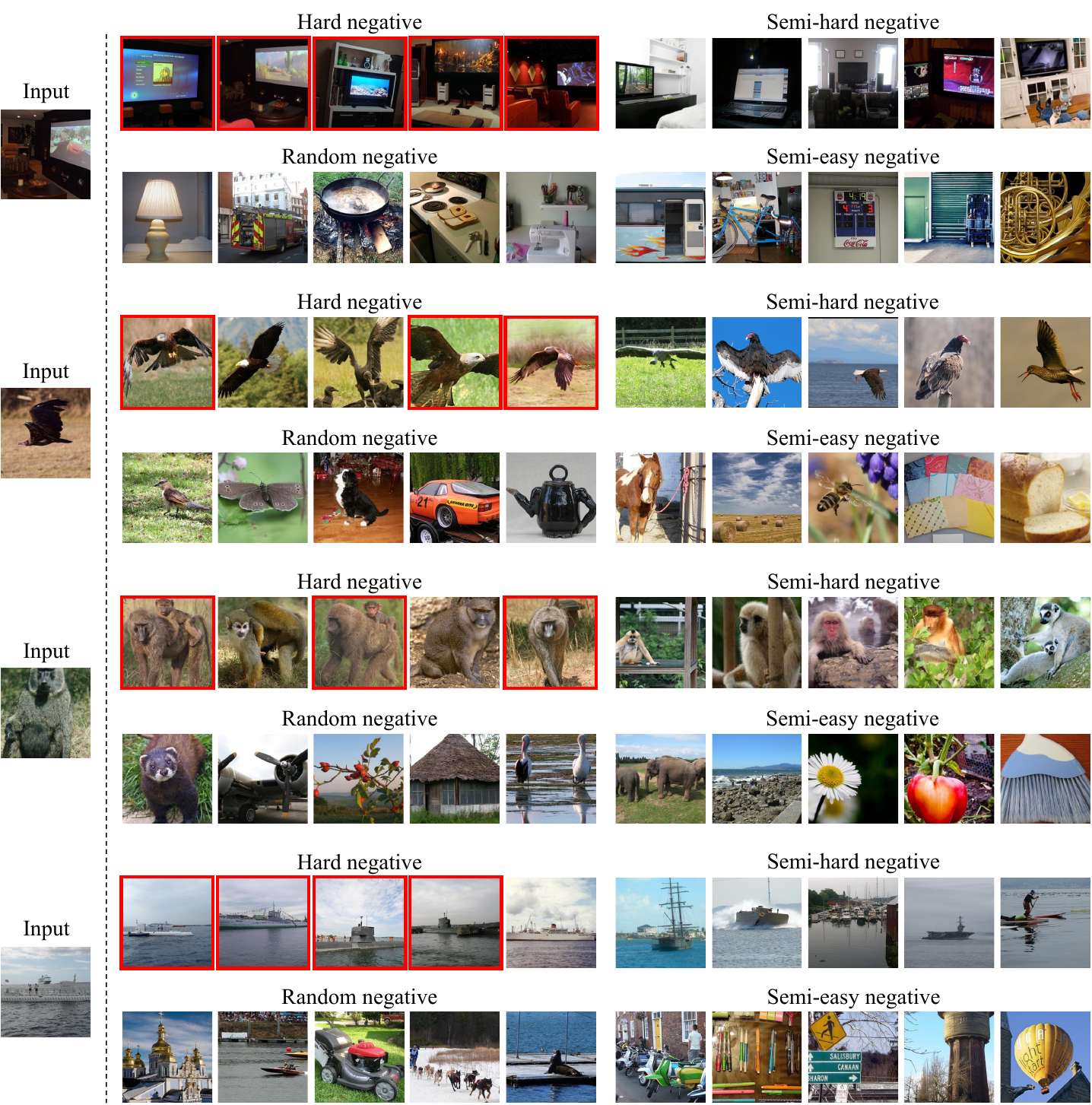}
  \caption{\textbf{Visualization of some example negative samples on ImageNet.} We randomly select five negative samples with different sampling strategies defined in Sect.~\ref{subsec:sampling} (\ie, hard, semi-hard, random, and semi-easy) for each input from the training set. The negative samples framed in \textcolor{red}{red} are in the same category as the input, \ie, false negatives}
  \label{fig:vlz_sampling}
\end{figure*}

\noindent\textbf{Negative sample visualization.}
Fig.~\ref{fig:vlz_sampling} visualizes some example negative samples with different negative sampling strategies (\ie, hard, semi-hard, random, and semi-easy) using the learned instance features during training.
For ``hard negative'' sampling, we observe many false negatives, \ie, they actually have the same category as the input image. For ``random negative'' and ``semi-easy negative'' sampling, although no false negatives are observed, the sampled negatives are largely visually dissimilar to the input and are much easier to be distinguished. In contrast, ``semi-hard negative'' sampling reduces the false negative cases while maintaining the difficulty of sampled negatives at the same time.

\begin{figure*}[ht]
  \centering
  \includegraphics[width=\linewidth]{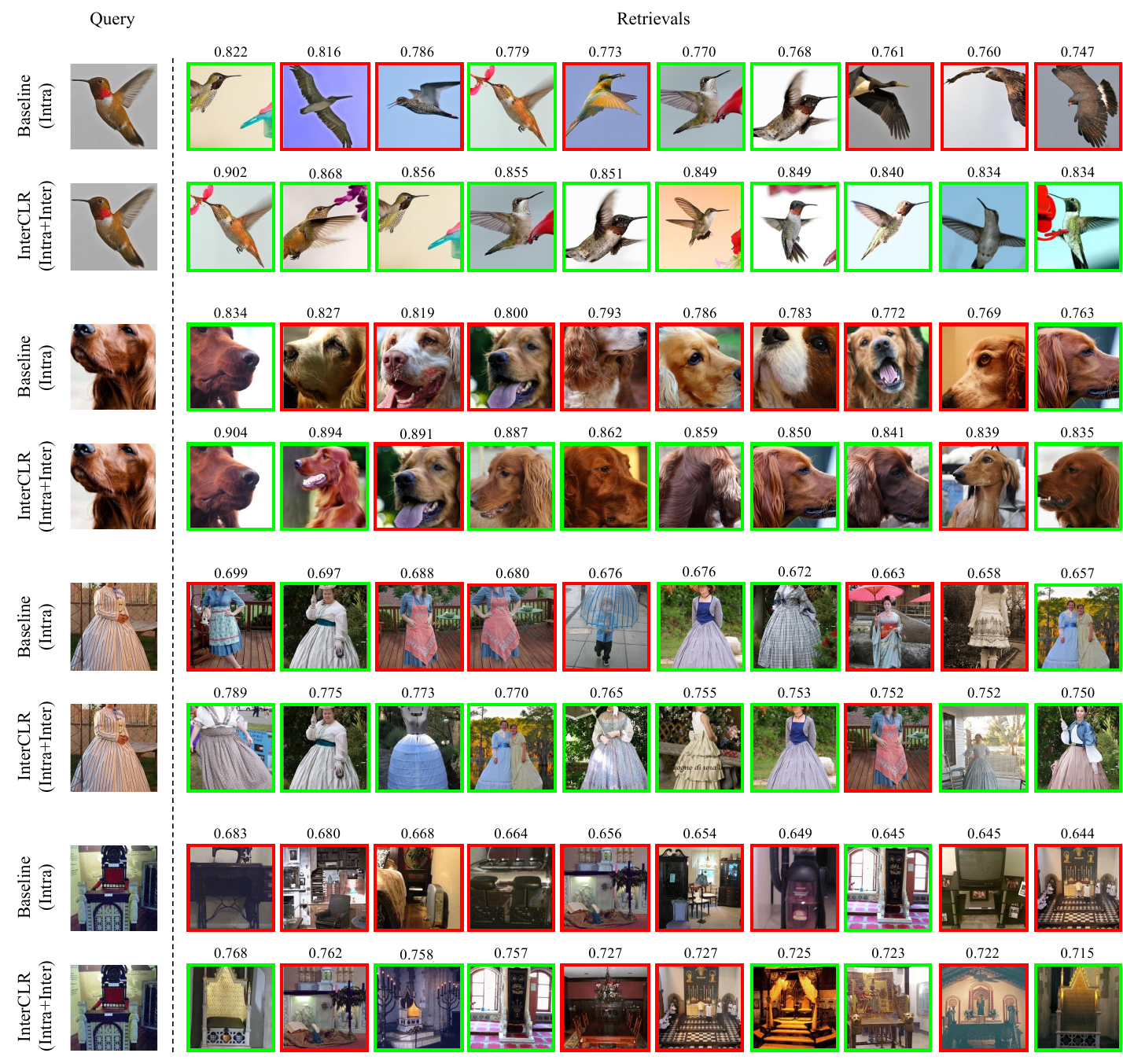}
  \caption{\textbf{Retrieval results of some example queries on ImageNet.} We compare InterCLR with its intra-image invariance learning baseline, \ie, Baseline (Intra). The left-most column are queries from the validation set, while the right columns show 10 nearest neighbors retrieved from the training set with the similarity measured by cosine similarity. The positive retrieved results are framed in \textcolor{green}{green}, while the negative retrieved results are framed in \textcolor{red}{red}. Number on the top of each retrieved sample denotes the cosine similarity with its corresponding query}
  \label{fig:knn_vlz}
\end{figure*}

\noindent\textbf{KNN visualization.}
To further understand the benefits of inter-image invariance learning, we visualize some top-10 nearest neighbors retrieved with cosine similarity in the embedding space using the features learned by InterCLR.
As shown in Fig.~\ref{fig:knn_vlz}, compared with its intra-image invariance learning baseline, \ie, Baseline (Intra), InterCLR (Intra+Inter) retrieves more correct images.
Besides, InterCLR (Intra+Inter) also achieves higher cosine similarity with the queries than its intra-image invariance learning counterpart.
We observe this not only for the same correctly retrieved samples, but also for the whole 10 retrieved nearest neighbors: even the 10th nearest neighbor retrieved by InterCLR (Intra+Inter) obtains higher cosine similarity than the 1st nearest neighbor retrieved by the baseline.
The aforementioned gap is due to the inherent limitation of intra-image invariance learning: \emph{only encouraging the similarity of different augmented views of the same image while discouraging the similarity of different images even though they should belong to the same semantic class.}
This further demonstrates the benefits of inter-image invariance brought by InterCLR.

\section{Conclusion}
\label{sec:conclusion}

In this work, we have investigated inter-image invariance learning from different perspectives and shown the effect of different design choices, \wrt pseudo-label maintenance, sampling strategy, and decision boundary design.
By combining our observations, we introduced a unified and generic framework, InterCLR, for unsupervised intra- and inter-image invariance learning.
With this framework, we consistently improve over existing state-of-the-art intra-image invariance learning methods on multiple standard benchmarks.
We hope our empirical study can provide useful baselines and experience for future research.

\noindent\textbf{Limitations.}
This study mainly targets at a low-resource pre-training regime, \ie, a batch size of 256 for 200 epochs. We have shown that InterCLR can also benefit from longer pre-training epochs as many prior works do in the paper. Pre-training with a larger batch size may further improve the performance. However, it usually comes at the cost of huge computational resources that are inaccessible to many researchers, which is not the core of this paper. We wish to highlight that our work wins with its merit of comprehensively revealing how to deal with the inaccurate nature of inter-image invariance learning from different perspectives instead of a large compute. It should be well noted that our 200-epoch InterCLR has already performed on par or even better than many prior large-batch long-epoch results attained with a much larger compute (\eg, SimCLR~\citep{chen2020simple}, SwAV~\citep{caron2020unsupervised}, BYOL~\citep{grill2020bootstrap}, and Barlow Twins~\citep{zbontar2021barlow}) on various downstream tasks. This demonstrates the appealing performance efficiency of InterCLR, \ie, one can get higher performance after training for fewer epochs, which is indeed important considering the high compute cost of existing unsupervised learning methods. Meanwhile, we choose representative intra-image invariance learning baselines and transfer learning benchmarks to examine the generality of our framework. More baselines and benchmarks can be further studied. We leave these explorations to future work.

\noindent\textbf{Data availability statements.}
The datasets that support the findings of this study are all publicly available for the research purpose.

\begin{acknowledgements}
This study is supported under the RIE2020 Industry Alignment Fund – Industry Collaboration Projects (IAF-ICP) Funding Initiative, as well as cash and in-kind contribution from the industry partner(s). The project is also supported by Singapore MOE AcRF Tier 2 (T2EP20120-0001), the Data Science and Artificial Intelligence Research Center at Nanyang Technological University.
\end{acknowledgements}

{\small
\bibliographystyle{spbasic}
\bibliography{bib}
}

\end{sloppypar}
\end{document}